\documentclass[letterpaper]{article}
\usepackage{aaai19}
\usepackage{times}
\usepackage{helvet}
\usepackage{courier}
\usepackage{url}
\usepackage{graphicx}
\usepackage[utf8]{inputenc} 
\usepackage{amsfonts}       
\usepackage{microtype}      

\usepackage[ruled,vlined]{algorithm2e}	
\usepackage[caption=false]{subfig}
\usepackage{amsmath}
\usepackage{mathtools}
\usepackage{tabu}

\title{Understanding Learned Models by Identifying\\Important Features at the Right Resolution}
\author{Kyubin Lee\textsuperscript{*}\\ Clinical Genomics Analysis Branch\\ National Cancer Center\\ Republic of Korea
 \And Akshay Sood\textsuperscript{*}\\ Dept. of Computer Sciences\\ Dept. of Biostatistics \& Medical Informatics\\ University of Wisconsin-Madison\\ \\ \small{\textsuperscript{*}These authors contributed equally to this work.}
\And Mark Craven\\ Dept. of Biostatistics \& Medical Informatics\\ Dept. of Computer Sciences\\ University of Wisconsin-Madison}

\begin{document}
\maketitle
\begin{abstract}
In many application domains, it is important to characterize how complex
learned models make their decisions across the distribution of
instances.  One way to do this is to identify the features and interactions
among them that contribute to a model's predictive accuracy.
We present a model-agnostic approach to this task that
makes the following specific contributions.  Our approach (i) tests
feature groups, in addition to base features, and tries to determine
the level of resolution at which important features can be determined,
(ii) uses hypothesis testing to rigorously assess the effect of each feature on
the model's loss, (iii) employs a hierarchical approach to control the
false discovery rate when testing feature groups and individual base
features for importance, and (iv) uses hypothesis testing to identify
important interactions among features and feature groups.  We evaluate
our approach by analyzing random forest and LSTM neural network models
learned in two challenging biomedical applications.
\end{abstract}

\section{Introduction \label{intro.sec}}

In many application domains, it is important to be able to inspect, probe, and understand models learned by machine-learning systems.
There are several principal reasons why it might be critical to understand how learned models make their decisions: (i) \textit{Trust}: end users and other stakeholders need to trust the models' decisions and understand the basis for them in order for the models to be accepted and employed; (ii) \textit{Model development}: to help improve the predictive performance of models, interpretable descriptions can aid in selecting among models, detecting and avoiding overfitting, and gaining insight into differences among input representations; (iii) \textit{Discovery}: our knowledge of a problem domain can be augmented by identifying previously unrecognized salient features and relationships that models have learned.

In such application domains, there is a strong incentive to use a
learning method that directly learns interpretable models, such as
logistic regression or a generalized additive model \cite{lou.kdd13}.
However, there is often tension between the desiderata of model
comprehensibility and predictive performance.  It may be the case that
the machine-learning approaches that provide the best predictive
performance in a given domain learn models that are highly challenging
to inspect and understand.  For this reason, a number of approaches
have been developed for gaining insight into complex learned models
such as random forests and deep neural networks.

Methods for gaining comprehensible descriptions of learned models can
be divided into two broad categories.  The first category, which is
referred to as \textit{prediction interpretability} encompasses
methods that lend insight into learned models by locally explaining the
decisions they make for individual instances
\cite{alvarez-melis.emnlp17,fong.iccv17,koh.icml17,lei.emnlp16,leino.arxiv18,ribeiro.kdd16,ribeiro.aaai18}.  The second
category, referred to as \textit{model interpretability}, refers to methods that aim
to provide characterizations of how models make decisions across the
distribution of instances.  Some methods in this category are tailored
to specific types of models
\cite{bau.cvpr17,bojarski.arxiv17,hara.aistat18,karpathy.iclr16},
whereas others are agnostic to the model type
\cite{craven.nips8,ribeiro.kdd16,ribeiro.aaai18}.

Here we present a model-agnostic approach that is focused on gaining
model interpretability.  The crux of our approach is to identify
important features, groups of features, and interactions among them.
The prior research that is most closely related to ours includes
methods that aim to provide model interpretability by identifying
important features through perturbations of input
\cite{breiman.mlj01,friedman.annals01,li.corr16}. There are also
methods that identify important features, but which are not model
agnostic \cite{epifanio.bmcbio17,fabris.bioinformatics18}.  The
specific contributions of our approach are the following.  First, it
is well suited to tasks with large, structured feature spaces.  In
such applications, the base features that are used as input to the
model might not provide the best level of resolution for understanding
or characterizing what is important to the learned model.  Our
approach tests feature groups, in addition to base features, and tries
to determine the level of resolution at which we can determine the
important features.  Second, we go beyond just ranking features
according to their importance, and instead use hypothesis testing to
assess the effect of each feature on the model's loss.  Given the
potentially large number of hypothesis tests that must be done, we use
a hierarchical approach to control the false discovery rate when
testing feature groups and base features for importance.  Third, we
propose a method based on hypothesis testing to identify important
interactions among base features and feature groups.

We evaluate our approach by analyzing random forest and LSTM neural
network models learned in two application domains: identifying viral
genotype-to-disease-phenotype associations, and predicting asthma
exacerbations from electronic health records (EHRs).  Additionally, we
validate our approach using synthetic data sets in which we know which
features and groups are truly important.



\section{Methods \label{methods.sec}}

In this section, we describe the key elements of the model-agnostic
approach we have developed for characterizing learned models. The source code for our methods is available at \url{https://github.com/Craven-Biostat-Lab/mihifepe}.

\subsection{Identifying Important Features via Perturbation \label{important.sec}}

As shown in Algorithm~\ref{feature_importance.alg}, a general
approach to identifying important features in a learned model is to
measure how the output of the model, or its loss, varies when
individual features in a given set of instances are perturbed in some
way.  Breiman \shortcite{breiman.mlj01} proposed an approach based on this
idea as a way to characterize learned random forest models, and Friedman proposed a similar approach for generating \textit{partial dependency plots} \cite{friedman.annals01,friedman.annals08}.  In
Breiman's method, the perturbation is done by permuting the values of
the given feature across a set of instances.  However, the approach
can be generalized to other perturbations, including feature
``erasure'' \cite{li.corr16}, flipping binary features, or replacing
features with ``background'' values.

\SetKw{Given}{given:}

\begin{algorithm}
	\SetKwInOut{Input}{input}\SetKwInOut{Output}{output}
	\SetKwProg{Fn}{function}{\string:}{}

	\DontPrintSemicolon
	\Input{
                learned model $h$, feature set $F$, test set $T = \{ \left( \mathbf{x}^{(1)}, y^{(1)} \right) \ldots \left( \mathbf{x}^{(m)}, y^{(m)} \right) \}$
	}\;
	\Output{
                set $\{ \left( j, v_j \right) | j \in F \}$ summarizing the effect $v_j$ on loss~$L$ when perturbing each feature $j$ 
	}\;
	\ForEach{\textup{feature} $j \in F$}{
		\ForEach{\textup{instance} $\left( \mathbf{x}^{(i)}, y^{(i)} \right)$ \textup{in} $T$}{
			let $\Delta \mathbf{x}_j^{(i)}$ represent $\mathbf{x}^{(i)}$ with feature $j$ perturbed in some way\;
			compare loss $L \left[  y^{(i)}, h\left( \mathbf{x}^{(i)} \right) \right]$ to  $L \left[  y^{(i)}, h\left( \Delta \mathbf{x}_j^{(i)} \right) \right]$\;
		}
		calculate summary statistic $v_j$ characterizing the effect of perturbing feature $j$ on $L$
	}
	\caption{General approach to identifying important features via perturbation\label{feature_importance.alg}}
\end{algorithm}

A key extension of this idea in our approach is that it uses
hypothesis testing to determine whether a given feature has a
generally consistent effect on the model's loss across the
distribution of instances.  We do this using held-aside test instances
so that our importance assessment measures whether a feature truly
impacts a model's predictive accuracy.  In the results presented here,
we use the Wilcoxon matched-pairs signed-rank test to assess the null
hypothesis that the median difference between pairs:
\begin{equation}
L\Big[ y^{(i)}, h\left( \mathbf{x}^{(i)} \right) \Big] - 
\frac{1}{P} \sum_{p=1}^{P} L\Big[ y^{(i)}, h\left( \Delta \mathbf{x}_j^{(i,p)} \right) \Big]
\label{basic-test}
\end{equation}
is zero.
Here $\Delta \mathbf{x}_j^{(i,p)}$ is defined as $\mathbf{x}^{(i)}$ with feature $j$ perturbed on the $p^{\text{th}}$ permutation.  For perturbations that do not involve randomness, such as erasure, $P = 1$ and $\mathbf{x}_j^{(i,1)}$ denotes the single perturbation that can be done to feature $j$. 

We use the Wilcoxon test
in place of a paired $t$-test due to significant non-normality in the
changes to loss introduced by feature perturbations. Here, we use the
one-tailed version of the test, corresponding to the median difference
being \textit{greater} than zero, in order to focus on features that
provide predictive value to the model.  Alternatively, we could use a two-tailed test
to also detect features whose perturbation \textit{decreases} loss, thereby indicating overfitting.

\subsection{Considering Feature Groups}

The approach described in Algorithm~\ref{feature_importance.alg} is
typically applied to the set of features that are used as input to the
model, which we refer to as \textit{base features}. We argue that, in
many domains, characterizing the importance of base features may not
be the right level of resolution for gaining a thorough understanding
of a learned model.  In some domains, there may be a large
number of features that are important to the model, and it may be
difficult to discern which high-level factors are most important for
the model's predictions unless groupings of related features are
considered.  For example models that perform risk assessment from
electronic health records often have thousands of base features
representing distinct diagnoses.  Our understanding of such a model is
likely to be aided by analyzing the importance of groups of related
diagnoses, or even the entire set of diagnoses, in addition to very specific ones.
Moreover, it might be the case that few, if any, individual base
features show a statistically significant change to the model's loss
when perturbed, or the effect sizes of these changes to the loss are
small.  In such cases, we can potentially detect statistical
significance and larger effect sizes by considering groups of related
features.

In contrast to assessing feature importance only at the level of base
features, our approach also assesses the importance of \textit{feature
groups}.  We assume that we are given a hierarchy in which internal
nodes represent groups of features, and leaf nodes represent base
features.  We can then apply Algorithm~\ref{feature_importance.alg} to both base features and feature groups in order to determine which are important.

In some application domains, such as risk assessment from EHRs,
there are standard ontologies which can be used to
define the hierarchy of feature groups.  For example, the
International Classification of Diseases (ICD) and the Clinical
Classifications Software (CCS) both define hierarchies of semantically
related groups of diagnoses and procedures.  In a risk-assessment
application, the base features might represent the occurrence of
specific recorded diagnoses in a given patient's EHR, such as
\textsf{reflux esophagitis} (ICD-9 code 530.11) or \textsf{acute
  esophagitis} (ICD-9 530.12).  We could test the importance of such
features by erasing all occurrences of the given diagnosis from
patients' records and measuring the resulting loss.  Moreover, we might test the importance of the
feature groups \textsf{esophagitis} (ICD-9 530.1), which has five
children diagnoses including the two listed above, or \textsf{diseases
  of the esophagus} (ICD-9 530), which has 28 descendant diagnoses. To
test a feature group, we could erase all recorded diagnosis that are
encompassed by the group.

In other application domains, the feature groups might be derived from
data.  For example, in our viral genotype-to-phenotype task, we calculate feature
groups using a hierarchical clustering method.  Our
base features are \textit{haplotype blocks}, which
are variable-sized regions of the genome that have been inherited as a
unit from one of two parental virus strains.  Our feature groups
consist of sets of neighboring haplotype blocks (i.e., larger regions
of the viral genome).

In a natural language domain, we might define feature groups on the
basis of syntactic or semantic categories.  
In an image classification domain, the base features might correspond to pixels
and we might define feature groups to represent superpixels or objects as feature groups. Perturbations could
involve replacing a region with a constant value, injecting
noise, or blurring \cite{fong.iccv17}.

In domains with temporal or sequential input, feature groups could
represent sets of features with restrictions based on their occurrence
in time/sequence.  For example, in a clinical risk-assessment domain
we might define feature groups representing occurrences of diagnoses
restricted to certain time windows, such as \textsf{diseases of the
esophagus} \textit{within the past year},
or \textsf{esophagitis} \textit{when patient's age} $> 50$.

In contrast to approaches for hierarchical feature
selection \cite{wan.aireview18}, the hierarchies used by our approach
do not necessarily represent \textit{is-a} or
\textit{generalization-specialization} relationships.  Each internal
node needs only to group features that are related in some
 meaningful way (e.g., neighboring regions of a genome).
Moreover, our approach is not focused on feature selection per se, but
instead on characterizing which feature groups are important in a
given learned model.

\subsection{Controlling the False Discovery Rate}

Given a hierarchy over the features, we can compute the effect of
perturbing each base feature and each feature group using
Algorithm~\ref{feature_importance.alg} across a given set of
instances. We treat each node in the hierarchy as representing the
null hypothesis that perturbing the corresponding feature group does
not have a significant effect on the loss function, in the sense that
the median of the differences computed using Formula~(\ref{basic-test})
is zero. A hypothesis
is rejected if this median difference is statistically significantly
different from zero, and a hypothesis is tested only if its
parent hypothesis has been rejected.

However, there is a notable multiple-comparisons problem
due to the potentially large number of hypotheses tested.
For instance, there are 8,740 hypotheses to be tested (counting both base features and feature groups) in the asthma exacerbation prediction task that we address.
Moreover, when adjusting for multiple comparisons, we need
to take into account the hierarchical organization of the hypotheses
being tested.  We address this issue by using the hierarchical false
discovery rate (FDR) control methodology developed by
Yekutieli \shortcite{yekutieli.jasa08} as described in Algorithm
~\ref{hierarchical_fdr.alg}.

This algorithm uses a recursive procedure to consider a hierarchical set
of hypotheses, which in our case consist of feature groups to be
tested.  If the null hypothesis is rejected for a given node in the
hierarchy (i.e., we determine that a feature group is important), then
the children of that node are tested using the Benjamini-Hochberg
method \cite{benjamini.jrss95} to control false discoveries.  Otherwise, the descendants of the
given node are not tested. The algorithm returns a subtree
representing the set of feature groups and base features for which the
null hypothesis was rejected.

Using this algorithm, we can identify the set of feature groups and
base features that have a significant effect on a model's loss while
controlling the rate of false discoveries in this set.  Of particular
interest is the set of \textit{outer} nodes: those nodes for which we
reject the null hypotheses (i.e., determine that they are important)
that have no children for which we reject the null hypotheses.  These
nodes represent the finest level of resolution at which we can
determine the importance of features and feature groups.

The key assumptions made by this approach, which are reasonable in our context, are that
(i) if a given feature significantly affects the loss when perturbed, a
group of features containing this feature will also significantly
affect the loss when perturbed, (ii) the $p$-values for siblings are
independently distributed, and (iii) $p$-values for true null hypotheses
are uniformly distributed in [0,1].

\begin{algorithm}[ht]
	\SetKwInOut{Input}{input}\SetKwInOut{Output}{output}
	\SetKwProg{Fn}{function}{\string:}{}
	\SetKwFunction{HierarchicalFDR}{HierarchicalFDR}%
	\SetKwFunction{MakeSingletonSet}{MakeSingletonSet}
	\SetKwFunction{Subtree}{Subtree}

	\DontPrintSemicolon
	\Input{
		Tree $\mathbb{T}$ of hypotheses to be tested along with their associated $p$-values, significance level $q$
	}\;
	\Output{
		A subtree $\mathbb{S}$ of $\mathbb{T}$ corresponding to hypotheses rejected while controlling FDR at significance level $q$
	}\;
	\Fn{ \HierarchicalFDR(node)} {
                // node has already been rejected\;
		\textit{rejectedSet} = \{ \textit{node} \}\;

		\If { node \textup{is not leaf} } {
			let $P_{(1)} \leq \ldots \leq P_{(k)}$ be the set of ordered $p$-values of \textit{node}.children\;
                        // Apply Benjamini-Hochberg procedure to children\;
			let $r = \max\{i: P_{(i)} \leq \frac{i \times q}{k}\}$\;

			\If {$r > 0$} {
				\textit{rejectedChildren} $=$ set of $r$ hypotheses corresponding to $P_{(1)} \leq \ldots \leq P_{(r)}$
			}
			\ForEach{ child $\in$ rejectedChildren }{
				\textit{rejectedSet} $=$ \textit{rejectedSet} $\cup$ \HierarchicalFDR(\textit{child})
			}
		}
		return \textit{rejectedSet}\;
	}\;
	\Begin {
		\eIf { $\mathbb{T}$.root.pvalue $>$ $q$ } {
			$\mathbb{S}$ = empty tree
		} {
			$\mathbb{S}$ = \HierarchicalFDR($\mathbb{T}$.\textit{root})
		}
	}
	\caption{Using hierarchical FDR control to identify important features\label{hierarchical_fdr.alg}}
\end{algorithm}

\subsection{Identifying Important Interactions}

In addition to identifying individual base features and feature groups
that are important, we would also like to identify interactions among
them that a given model has determined as important.  Here we
consider cases in which the model outputs a scalar value.  For this analysis, we do not treat a given model completely as a black box, but instead assume that we
know the transfer function that produces the model's outputs.  Let
$g(\mathbf{x}^{(i)})$ denote the function that maps $\mathbf{x}^{(i)}$
to the value that is input to the transfer function $f( \cdot )$, and $h \left( \mathbf{x}^{(i)} \right) = 
f \left( g(\mathbf{x}^{(i)}) \right)$ indicate the output of the model.  For
example, $f( \cdot )$ might be a logistic activation function in a neural
network for a binary classification task, in which case $g( \cdot )$ would
represent the part of the network that maps from $\mathbf{x}^{(i)}$ to the net
input of the logistic function.  Or in a random forest trained for a
regression task, $f( \cdot )$ would represent the identity
function, and $g( \cdot )$ would represent the average of the
values predicted by the individual trees in the forest.

Our notion of an interaction among features is based on the concept of additivity.
We define an interaction between feature $j$ and feature $k$ to mean that changes in $g( \cdot )$  when we perturb both features are non-additive (for some instances):
\begin{multline}
\left[ g\left( \Delta\mathbf{x}_j^{(i)} \right) - g\left( \mathbf{x}^{(i)} \right)  \right] + \left[ g\left( \Delta\mathbf{x}_k^{(i)} \right) - g\left( \mathbf{x}^{(i)} \right) \right] \not\approx \\ \left[ g\left( \Delta\mathbf{x}_{j \land k}^{(i)} \right) - g\left( \mathbf{x}^{(i)} \right) \right]
\label{interaction.outputs.eqn}
\end{multline}
where $\Delta\mathbf{x}_{j \land k}^{(i)}$ denotes instance $\mathbf{x}^{(i)}$ with feature $j$ and feature $k$ jointly perturbed.

To identify interactions that are important, we use hypothesis testing
to assess whether a candidate interaction exhibits nonadditivity.  We
can do this by considering the median difference between pairs formed
by the two sides of the inequality above.  In the results presented
here, we use the Wilcoxon matched-pairs signed-rank test to assess the
null hypothesis that the median difference between the pairs is zero.
This approach to testing interactions can be applied to base features, feature groups, and mixtures thereof.

Alternatively, we can consider whether a candidate interaction exhibits nonadditivity
 which has a generally consistent effect on
the model's loss across the distribution of instances. We can do this by
assessing the difference between
pairs:
\begin{multline}
	L \left[  y^{(i)}, f\left( g \big( \Delta\mathbf{x}_{j \land k}^{(i)} \big) \right) \right] - \\
        L \left[  y^{(i)}, f\left( g \big( \mathbf{x}^{(i)} \big) + \Delta g \big( \Delta \mathbf{x}_{j}^{(i)} \big) + \Delta g \big( \Delta \mathbf{x}_{k}^{(i)} \big) \right) \right]
\end{multline}
where $\Delta g \big( \Delta \mathbf{x}_{j}^{(i)} \big)$ is defined
as $\left[ g \big( \Delta \mathbf{x}_{j}^{(i)} \big) -
g \big( \mathbf{x}^{(i)} \big) \right]$ (i.e., the change in
$g(\mathbf{x}^{(i)})$ that results from perturbing feature $j$). 
However, the null distribution may not be as straightforward to work with in
this case because, depending on the loss function, the difference in variances of the inner terms on each side may lead to the loss terms having different means.

A related approach that can be used to detect interactions is the $H^2$ statistic \cite{friedman.annals08} which is based on partial dependency scores.


\section{Results \label{results.sec}}

In this section, we evaluate our approach by (i) assessing
its ability to detect important features and interactions while controlling FDR on synthetic
data sets, and (ii) applying it in two biomedical
domains in which it is essential to understand learned models.

\subsection{Evaluation on Synthetic Data Sets}

To verify that our approach is able to identify important features and interactions
while controlling the false discovery rate, we first evaluate it using
data sets for which we know the truly important features.
In this setting we can think of each model as approximating a
ground-truth function of the form:
\begin{equation}
y^{(i)} = \sum_{j \in I_L} \alpha_j x_j^{(i)} + \sum_{\mathclap{\substack{(j, k) \in I_I \\ j \neq k}}} \alpha_{jk} x_j^{(i)} x_k^{(i)}
\end{equation}
where $I_L$ and $I_I$ represent the subset of important linear and interaction terms respectively, and $\alpha_j$ and $\alpha_{jk}$ are corresponding coefficients that determine how the $j^{\text{th}}$ feature and $(j, k)^{\text{th}}$ interaction
contribute to the output. Note that a feature is considered important if belongs to $I_L$, or is a component of an interaction that belongs to $I_I$, or both. We represent a ``learned'' model using the
following form:
\begin{equation}
h \left( \mathbf{x}^{(i)} \right) = \sum_{j \in I_L} \alpha_j x_j^{(i)} + \sum_{\mathclap{\substack{(j, k) \in I_I \\ j \neq k}}} \alpha_{jk} x_j^{(i)} x_k^{(i)} + \gamma^{(i)}
\label{model.eqn}
\end{equation}
where $\gamma^{(i)} \sim N(0, \sigma^2)$ represents the deviation of
the model's output from the ground-truth function for some instance
$i$ in the feature space.  This formulation is intended to simulate
the situation in which a learned model provides a fairly accurate
representation of the underlying target function, but incorporates
irrelevant features and other deviations which have a small impact on the model's outputs.

We generate synthetic data sets by drawing feature vectors from a
given distribution, and then using Equation~\ref{model.eqn} to
determine $h(\mathbf{x}^{(i)})$ for each $\mathbf{x}^{(i)}$, and
similarly for each perturbation of $\mathbf{x}^{(i)}$.  Here we
present results in which our instance spaces have 500 binary features,
and each underlying ground truth function has 50 important features and 50 important interactions selected from among these, with coefficients $\alpha_j \sim U(0, 1)\;\;\forall j \in I_L$ and $\alpha_{jk} \sim U(0, 1)\;\;\forall (j,k) \in I_I$. The feature vectors are
constructed by sampling each feature from an independent Bernoulli
distribution.  We define feature groups by creating a balanced binary hierarchy with features randomly assigned to leaf nodes and feature groups represented by internal nodes.
A feature group is considered important if it contains at least one important feature in its subtree.
We use Equation~\ref{basic-test} for hypothesis testing of the features, performing perturbations by erasure (i.e., setting the feature to zero in all instances), followed by the hierarchical FDR procedure (Algorithm~\ref{hierarchical_fdr.alg}) with $q = 0.05$.

To analyze interactions, we use the (base) features identified as important in the preceding analysis to construct a set of potential interactions to test. This allows us to prune the large search space of all possible interactions, albeit at the cost of decreased power. We then use Equation~\ref{interaction.outputs.eqn} to perform hypothesis testing of these interactions, and use the Benjamini-Hochberg procedure \shortcite{benjamini.jrss95} to control FDR among this set.
 
\begin{table}[ht]
\centering
\subfloat[]{
\centering
\begin{tabu} to 0.95\linewidth {|X[0.7r]|X[c]X[c]|X[c]X[c]|}
\hline
\multicolumn{1}{|l|}{} & \multicolumn{2}{c|}{features} & \multicolumn{2}{c|}{interactions} \\
$m$ & FDR & power & FDR & power \\ \hline
32 & 0.019 & 0.722 & 0.046 & 0.132 \\
64 & 0.024 & 0.800 & 0.014 & 0.370 \\
128 & 0.026 & 0.850 & 0.030 & 0.543 \\
256 & 0.029 & 0.895 & 0.035 & 0.682 \\
512 & 0.036 & 0.919 & 0.040 & 0.777 \\
1024 & 0.035 & 0.936 & 0.039 & 0.840 \\
2048 & 0.029 & 0.948 & 0.048 & 0.877 \\
4096 & 0.029 & 0.960 & 0.045 & 0.913 \\
8192 & 0.033 & 0.967 & 0.046 & 0.935 \\
16384 & 0.032 & 0.975 & 0.039 & 0.949 \\ \hline
\end{tabu}
\label{Synthetic}}

\subfloat[]{
\centering
\begin{tabu} to 0.95\linewidth {|X[0.7c]|X[c]X[c]|X[c]X[c]|}
\hline
\multicolumn{1}{|l|}{} & \multicolumn{2}{c|}{features} & \multicolumn{2}{c|}{interactions} \\
$\sigma$ & FDR & power & FDR & power \\ \hline
0.00 & 0.000 & 0.999 & 0.000 & 0.991 \\
0.01 & 0.034 & 0.983 & 0.048 & 0.966 \\
0.02 & 0.034 & 0.982 & 0.047 & 0.964 \\
0.04 & 0.034 & 0.980 & 0.048 & 0.958 \\
0.08 & 0.034 & 0.974 & 0.048 & 0.945 \\
0.16 & 0.034 & 0.964 & 0.049 & 0.920 \\
0.32 & 0.034 & 0.938 & 0.048 & 0.866 \\
0.64 & 0.033 & 0.887 & 0.049 & 0.766 \\
1.28 & 0.033 & 0.770 & 0.050 & 0.564 \\ \hline
\end{tabu}
\label{Synthnoise}}

\caption{FDR and power on synthetic data sets as (a) the size of the test set $m$ is varied (b) $\sigma$ is varied.}
\end{table}

Table~\ref{Synthetic} shows the results of applying our method as 
 the number of instances in the ``test set'' is varied.
 The results in the table represent averages over 100 randomly generated models and datasets. For each test-set size,
we report both the mean power of the method (i.e., the fraction of
the truly important features and interactions that are identified as
such) and the mean false discovery rate (i.e., the fraction of putatively
important features and interactions that are not truly important).
The middle columns show FDR and power when determining important features and feature groups, and the rightmost columns
show FDR and power when determining important interactions.
Table~\ref{Synthnoise} shows the effect of varying $\sigma$ when sampling the $\gamma^{(i)}$ values for each learned model.  Here, the number of instances is 10,000.
The results in Tables~\ref{Synthetic} and~\ref{Synthnoise} indicate, not surprisingly, that the power of our method to detect truly important features and interactions increases with larger test sets, and decreases with larger values of $\sigma$.  Importantly, for all conditions, the FDR $\leq 0.05$ as expected with our approach.

The analyses of both features and interactions show similar trends. However, the mean power for discovering important interactions trails the mean power for discovering important features for any given test set size/noise level. This is because we only test an interaction if its constituent features have already been found to be important during the preceding feature analysis.

\subsection{Real Application Domains and Models}


The first real domain we consider is focused on identifying the genetic
components of Herpes simplex virus type~1 (HSV-1) that are responsible
for various dimensions of eye disease.  Here we analyze random-forest
models that have learned mappings from variations in viral genotypes
to three different eye disease phenotypes
\cite{kolb.plospathogens16,lee.jvirology16}.  Each instance
corresponds to a genetically distinct strain of the virus, and there are 65
recombinant strains generated by mixed infection of two parental
strains. We represent each genotype as a vector of 547 features, where
each feature corresponds to a \textit{haplotype block} which is
variable-sized regions of the genome that has been inherited as a unit
from one of the two parental virus strains.  The value of each binary
feature indicates from which parental strain the haplotype block was
inherited. The phenotypes (blepharitis, stromal keratitis, and
neovascularization) for each instance are numeric scores indicating
the disease severity resulting from infection in mice by a given
strain.  
The random-forest (RF) regression models had statistically
significant predictability for all three phenotypes and they
demonstrated better cross-validated predictive accuracy than penalized
linear regression models (Lasso and Ridge) for two of these three
phenotypes, and others we have assessed. The cross-validated $R^2$ values for the 
blepharitis, stromal keratitis, and
neovascularization models are 0.45, 0.56, and 0.48, respectively.
Each learned RF model
comprises 1,000 trees.

The second application domain we address is to predict asthma
exacerbations from electronic health records.  The data set consists
of information derived from EHRs for a cohort of 28,101 asthma
patients from the University of Wisconsin Health System over a five-year period.  The
information extracted from the EHRs includes demographic features and
time-stamped events corresponding to encounters with the healthcare
system.  These events include problem-list and other coded diagnoses,
procedures, medications, vitals, asthma control scores, and prior
exacerbations.  We also include features representing the time since
the last event, represented at multiple scales.

We learned long short-term memory (LSTM) neural networks
\cite{hochreiter.nc97} to predict whether a patient would experience
an exacerbation within the next 90 days or not given their clinical
history as represented in the EHR.  Deep recurrent neural networks (RNN)
have demonstrated state-of-the-art predictive accuracy in learning
models from healthcare data \cite{miotto.natsci16,pham.packdd16}.  Our
LSTM networks have a cell state of size 100 and a sigmoid output
layer. The coded diagnoses, problem diagnoses, and interventions
(procedures and medications) all comprise large vocabularies (6,533
for coded diagnoses, 4,398 for problem diagnoses, and 8,745 for
interventions) of which only a small subset is recorded at each
encounter.  Therefore, we first map event vectors for each of these sets
to an embedded space using Med2Vec \cite{choi.sigkdd16}, resulting in
shorter, dense fixed-length vectors.  Separate embeddings of size 200
were generated for each of these sets, which were then
concatenated, along with the other temporal features, to produce the
event representation at each timestamp in the record.  The ordered
sequence of events formed the input sequence for the LSTM.  The static
demographic features were provided as input at the output sigmoid
layer.  Using 10-fold cross-validation to assess the predictive
accuracy of the networks results in an area under the ROC curve
(AUROC) of 0.757.

\subsection{Feature Groups and Perturbations}

For the HSV-1 application, our feature hierarchy represents neighboring regions of the viral genome.
We compute the hierarchy 
using a constrained hierarchical clustering method applied to the base features, which represent haplotype blocks.
This clustering
method uses Hamming distance to compare columns (features) in our data matrix,
and a complete linkage function, such that every pair of features in a
given cluster is within a specified bit difference.  The agglomerative clustering operator groups features that are correlated 
(i.e., exhibit similar inheritance patterns) across the viral strains.
Since we want our
hierarchy to group \textit{neighboring} haplotype blocks that are correlated,
we constrain the
clustering method such that hierarchy adheres to the linear ordering
of the haplotype blocks with the HSV-1 genome.  Thus, the merging step
during clustering can be applied only to features or feature groups
that are adjacent to each other in the genome.  The resulting
hierarchy consists of 547 leaf nodes (base features) and 546 internal
nodes (feature groups).

The perturbations we use to interrogate models in this domain are
based on permutations. For a given feature or feature group, we
randomly shuffle and reassign the values for the feature (group) in the data matrix.  When doing such permutations for feature
groups, the values in the group for each instance are
treated as a unit, being shuffled and reassigned together.
We do this perturbation 500 times for each feature or feature group when assessing its importance.

We consider two hierarchies over features for the asthma exacerbation
prediction task.  We construct a top-level hierarchy representing our
broad categories of EHR-elicited features (diagnoses, demographics,
etc.).  The second hierarchy we use is the standard ICD-9 hierarchy of
diagnoses.  In this application, we use erasure perturbations which
involve zeroing out features or feature groups of interest, following the use of erasure by Li et al. \shortcite{li.corr16}.
For event-based features, the erasure operation we use removes all
occurrences of the feature from a patient's history.  For features
that are encoded in an embedded representation, the erasure operation
is applied to the patient's history and then the embedding of the
associated events is recomputed while keeping the embedding models the
same.

\begin{table*}
\centering
\resizebox{\linewidth}{!}{%
\begin{tabular}{|l|rrr|c|}
\hline
&\multicolumn{3}{c|}{HSV-1 genotype-phenotype association}                                   & asthma exacerbation  \\
& blepharitis & stromal keratitis & neovascularization & ICD-9 \\
\hline
total nodes (base features + feature groups) & 1,093 & 1,093 & 1,093 & 8,740 \\
nodes with unadjusted \textit{p} \textless{} 0.05 & 242 & 148 & 111 & 3,480 \\
nodes rejected at \textit{q} level \textless{} 0.05   & 107 & 110 & 80 & 3,179 \\
outer nodes & 40 & 36 & 24 & 2,120 \\
feature groups among outer nodes & 6 & 3 & 3 & 159 \\ \hline
\end{tabular}%
}
\caption{Summary of feature-importance hypothesis testing in both application domains.}
\label{Summary}
\end{table*}

\begin{figure*}
	\centering
	\includegraphics[width=\linewidth]{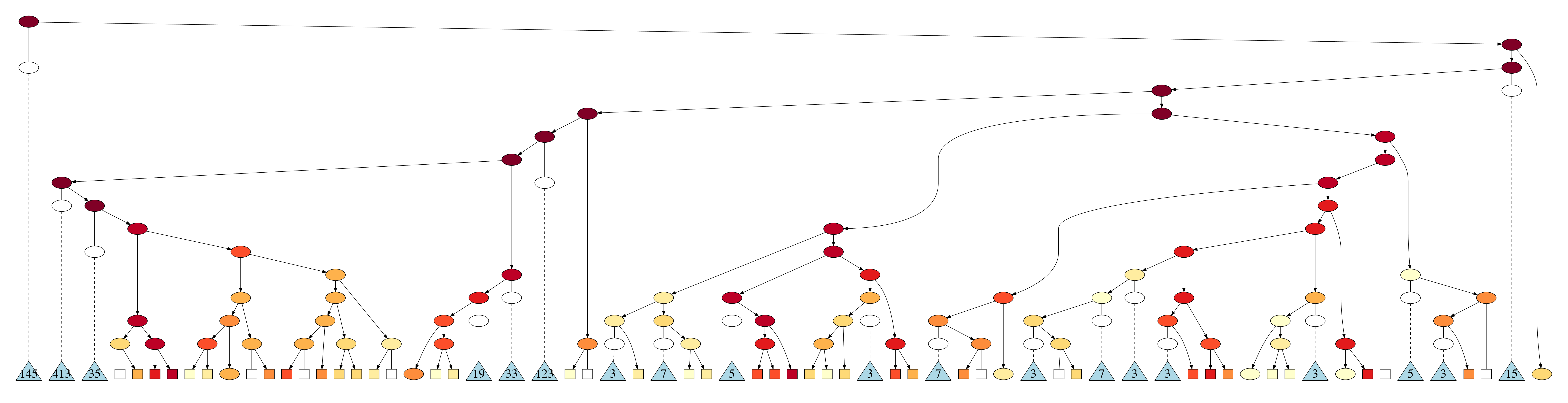}
	\caption{Feature importance analysis of the random forest model for blepharitis.  Ovals represent feature groups, squares depict base features, and triangles depict subtrees of the hierarchy that were not tested by the FDR procedure.  Color intensity indicates the magnitude of the associated $p$-value.  White nodes are those that were tested but did not survive the FDR procedure.}
	\label{fig:H_B}
\end{figure*}

\begin{figure*}
  \centering
  \subfloat[]{\includegraphics[width=\linewidth]{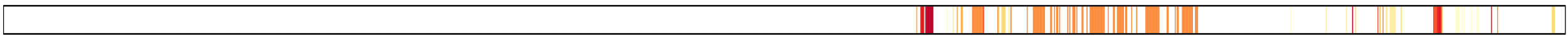}}

  \subfloat[]{\includegraphics[width=\linewidth]{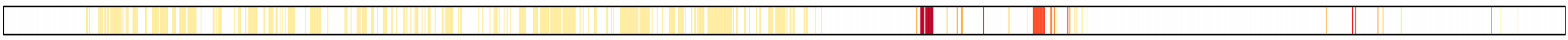}}

  \subfloat[]{\includegraphics[width=\linewidth]{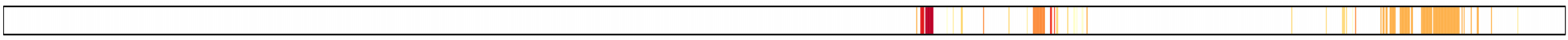}}

  \subfloat{\includegraphics[width=\linewidth]{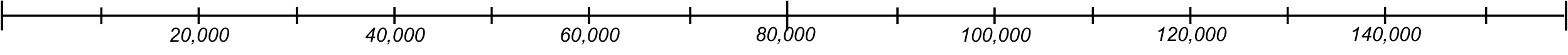}}

  \caption{Important features mapped to the HSV-1 genome coordinates for all three disease phenotypes: (a) blepharitis, (b) stromal keratitis, (c) neovascularization.  Color intensity indicates the magnitude of the associated importance $p$-value.}
  \label{fig:genomeMap}
\end{figure*}

\subsection{Identifying Important Features}

In this section, we determine which features and feature groups we can
identify as being important to our learned models in both application
domains when controlling the false discovery rate with $q = 0.05$.
Table~\ref{Summary} summarizes the results of our feature importance
analysis of models learned for four tasks in both domains.
The first row in the table indicates the number of base features and
feature groups that were assessed for each model.  The second row
indicates the number of base features and feature groups that have an
unadjusted $p$-value $< 0.05$ when doing significance testing as
described in the Methods Section.  The third row shows the
number of features that we ascertain are important after doing
hierarchical FDR control.  The last two rows indicate, among those
nodes surviving the FDR control, the number that are outer nodes, and
the number of outer nodes that correspond to feature groups.  Recall that outer
nodes refer to those that survive the FDR control but have no children
that do.

Figure~\ref{fig:H_B} provides a visual depiction of these results for
the blepharitis phenotype model.  
Among the 1,093 base
features and feature groups that were tested, we determine that 107
are important when controlling the FDR at $ q = 0.05$.  Moreover the set of
40 outer nodes represents the finest level of resolution at which we
can say that a viral genomic region is important to the phenotype.  In
the case of the blepharitis phenotype, six of the outer nodes are
feature groups which represent genomic regions that seem to be
associated with the phenotype but for which we cannot localize
precisely which base features are important.
Figure~\ref{fig:genomeMap} shows the identified important features for
all three disease phenotypes mapped to the genomic coordinates of the
virus.  Through the application of our approach to the learned RF
models, we are able to significantly narrow down the genetic
determinants of disease from a large number of candidate regions.
Several of these regions recapitulate what was previously known about
HSV-1 pathogenicity, and others indicate novel disease determinants \cite{kolb.plospathogens16,lee.jvirology16}.
Moreover, the results suggest a high degree of underlying causality
among the three disease phenotypes given the fact that there is
substantial overlap among the important regions identified.

\begin{figure*}[ht]
  \subfloat[]{
	\centering
	\includegraphics[width=0.95\linewidth]{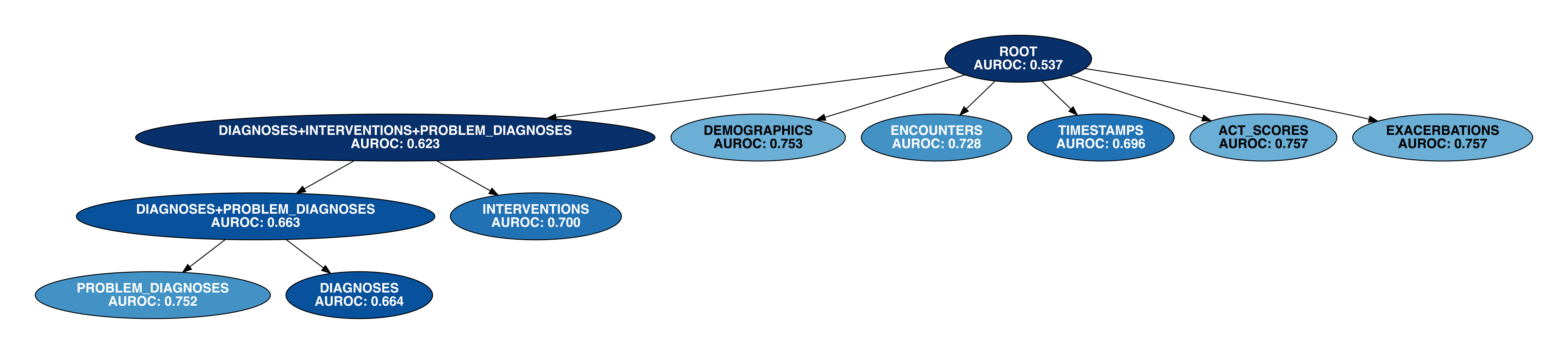}
	\label{fig:ehr_all}}

  \subfloat[]{
	\centering
	\includegraphics[width=0.95\linewidth]{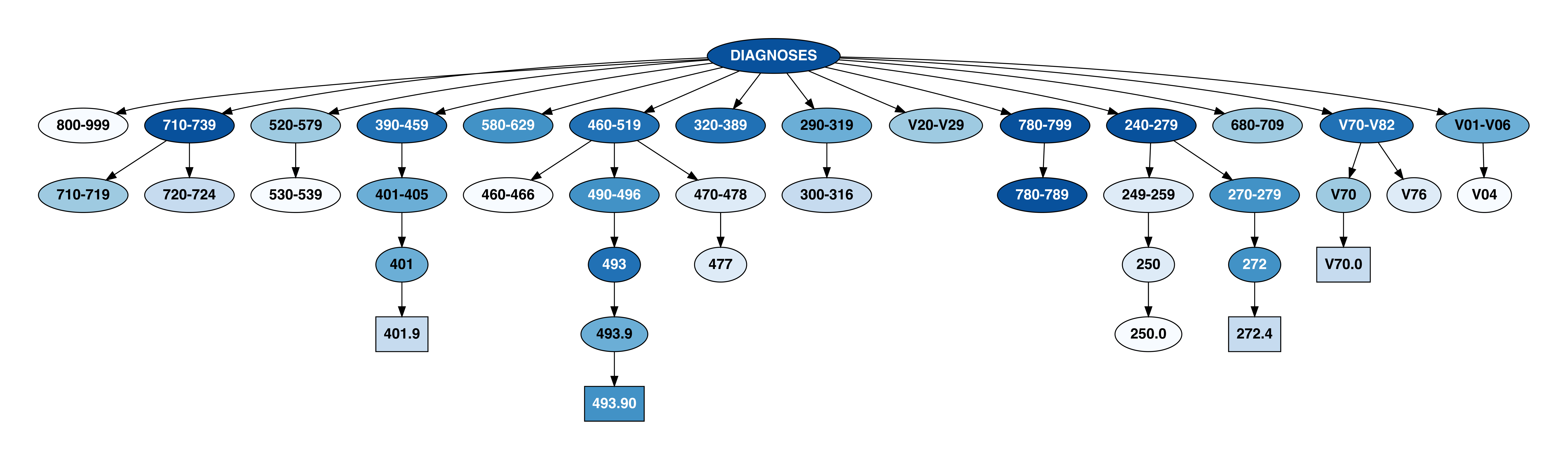}
	\label{fig:ehr_diagnoses}}
\caption{Feature importance analysis of the LSTM model for predicting asthma exacerbations.  Darker shades correspond to larger effect sizes (lower AUROCs when the feature groups are perturbed). (a) Importance analysis for highest level feature groups. (b) Importance analysis for feature groups representing the ICD-9 hierarchy of diagnoses. Note that the root node in panel (b) corresponds to the \textsf{DIAGNOSES} node in panel (a).}
\end{figure*}

Figure~\ref{fig:ehr_all} shows the results of our feature importance
analysis when applied to the highest level feature groups for the asthma-exacerbation model.  These
results suggest that the most informative feature groups are coded
diagnoses (\textsf{DIAGNOSES}), intervals between events
(\textsf{TIMESTAMPS}), and interventions (which combines medications
and procedures).  We note that even when all the features are erased
(\textsf{ROOT}), the model still has some predictive power with AUROC
= 0.537.  This is likely due to the fact that the number of encounters
in a patient's history is predictive.  Even when we erase all other
information, we leave the number of events in a patient's history
intact.
Figure~\ref{fig:ehr_diagnoses} partially depicts the results of our hierarchical FDR analysis on the diagnosis feature group.
These results are also summarized in Table~\ref{Summary}.
A large number of hypotheses are rejected at FDR control level $q = 0.05$, indicating that many features and feature groups have some predictive signal for this task.
Figure~\ref{fig:ehr_diagnoses} shows those nodes surviving the FDR control that have larger effect sizes.
The features identified by the analysis as important include those with known connections to asthma, such as the \textsf{respiratory diseases} subtree (460-519) terminating at \textsf{asthma} (493.90), and the \textsf{mental disorders} subtree (290-319) \cite{scott.ghp07}.
It also identifies important features with less well understood relationships to asthma, such as the \textsf{metabolic diseases} subtree (240-279).

\subsection{Identifying Important Interactions}

We also apply our approach for detecting important feature interactions to the models learned for HSV-1
genotype-phenotype associations.
We have not yet
developed an approach for effectively exploring the space of hypotheses
corresponding to interactions while controlling the FDR, so here we evaluate interactions
among two sets. First, we assess pairwise interactions between
all outer nodes that were determined as important in the individual feature
analysis. There are 780, 630, and 276 pairwise
interaction candidates to be tested by our approach
for blepharitis, stromal keratitis, and neovascularization, respectively.
After applying our hypothesis testing method for interactions to
the outer nodes, we use the Benjamini-Hochberg method \shortcite{benjamini.jrss95} to control the false discovery rate among this set.
Controlling FDR at 0.1,  there is only one surviving
interaction among the three phenotype models. For stromal keratitis, we
identified a significant interaction between two base features, where one
of the features is the one with the largest effect size among the outer nodes.

We also consider interactions among a set of nodes located at an intermediate
level in the hierarchy of features that survived FDR control in the individual feature
importance analysis.  We were able to detect several significant
interactions for the stromal keratitis phenotype. Among 435 candidate interactions tested,
three interactions were significant. 

\section{Conclusion \label{discussion.sec}}

We have presented a model-agnostic approach to understanding learned
models by identifying important features, and interactions among them, at various
level of resolution.  The key contributions of our approach are that
it employs hypothesis testing, along with hierarchical feature
groupings and a hierarchical-FDR control method, in order to
rigorously assess which features and groups of features have a
significant effect on a model's loss.  Moreover, we have also
presented an approach for testing important feature interactions.  We
have demonstrated and evaluated our approach in the context of two
biomedical domains.  In both domains, our method has lent insight into
complex learned models by determining important features and feature
groups.  Additionally, we have identified important interactions in
one of our HSV-1 models.

There are a number of directions we plan to explore in future work.
These include developing an effective approach for exploring the space
of candidate interactions,
assessing the importance of time-based feature groups in the context
of our asthma exacerbation model, and analyzing feature groups that
are organized into graphs that are not necessarily trees.

\section{Acknowledgments}

This work was funded by NIH grants U54 AI117924 and UL1 TR000427.


{\small \bibliography{main.bib}}
\bibliographystyle{aaai}
\end{document}


\maketitle

\begin{minipage}{\textwidth}
\centering
   \includegraphics[width=\textwidth]{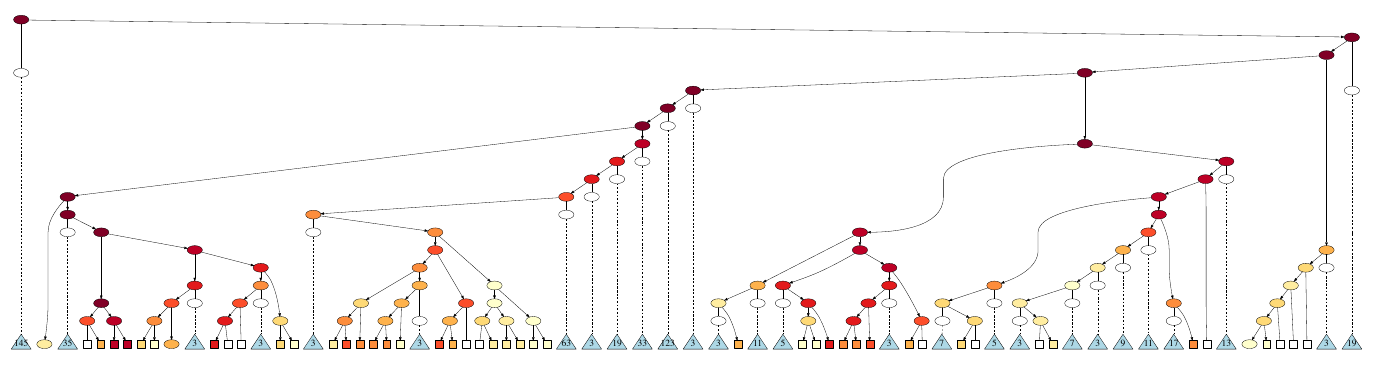}
	\captionof{figure}{Feature importance analysis of the random forest model for stromal keratitis. Ovals represent feature groups, squares depict base features, and triangles depict subtrees of the hierarchy that were not tested by the hierarchical FDR procedure. Color intensity indicates the magnitude of the associated $p$-value. White nodes are those that were tested but did not survive the FDR procedure.}
	\label{fig:stromal_keratitis}
\vspace*{20 pt}
	\includegraphics[width=\textwidth]{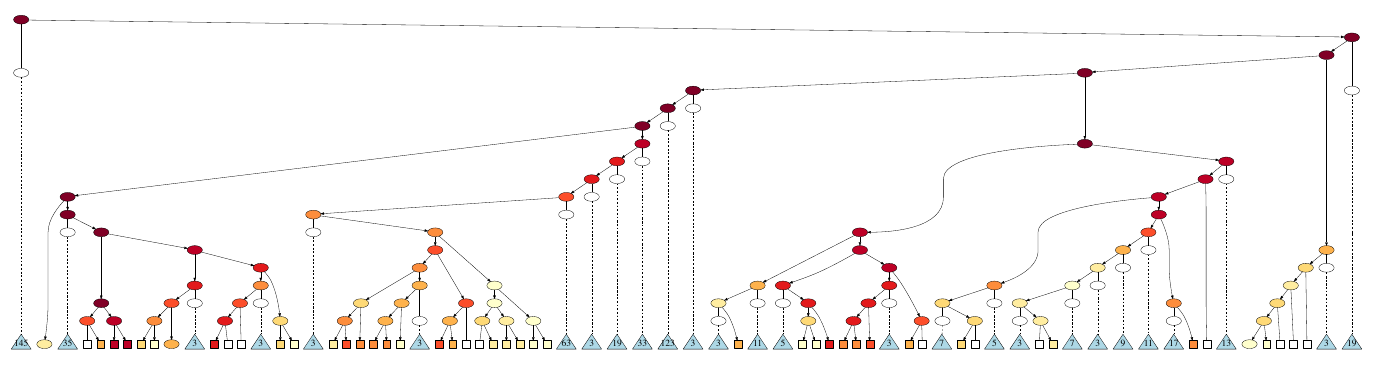}
	\captionof{figure}{Feature importance analysis of the random forest model for neovascularization.}
	\label{fig:neovascularization}
\end{minipage}

\begin{figure*}[t]
	\centering
	\includegraphics[width=\textwidth]{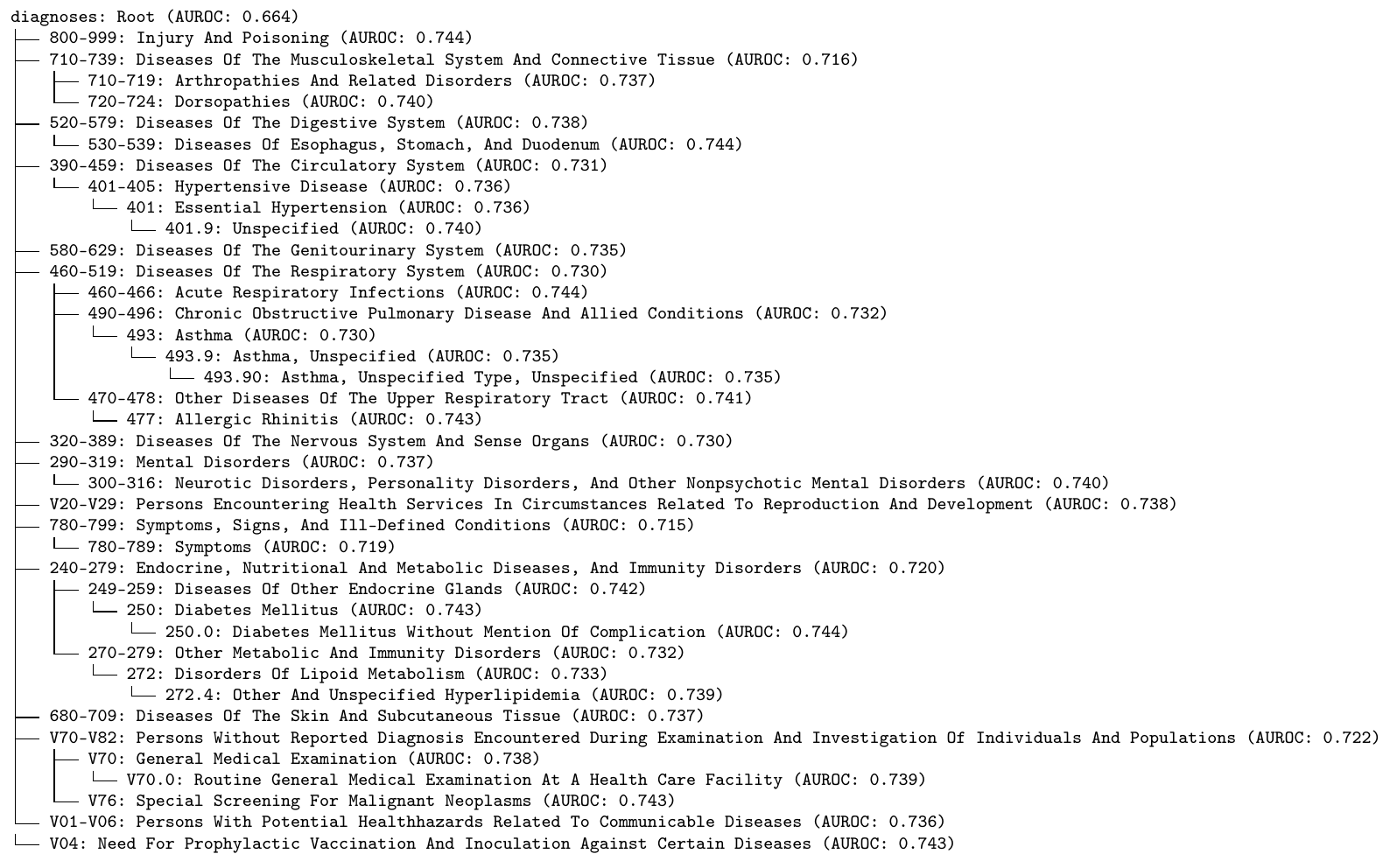}
	\caption{Descriptions of important ICD-9 features from Figure 3(b) from the main paper.}
	\label{fig:icd9_descriptions}
\end{figure*}